\documentclass[lettersize,journal]{IEEEtran}
\usepackage{amsmath,amsfonts}
\usepackage{algorithmic}
\usepackage{algorithm}
\usepackage{array}
\usepackage[caption=false,font=normalsize,labelfont=sf,textfont=sf]{subfig}
\usepackage{textcomp}
\usepackage{stfloats}
\usepackage{url}
\usepackage{verbatim}
\usepackage{graphicx}
\usepackage{cite}
\usepackage{color} 
\usepackage{caption}
\usepackage{multirow}
\usepackage[table,xcdraw]{xcolor} 
\usepackage[colorlinks,bookmarksopen,bookmarksnumbered,citecolor=blue, linkcolor=red, urlcolor=blue]{hyperref}
\begin{document}

\title{Learn2Talk: 3D Talking Face Learns from 2D Talking Face}

\author{Yixiang Zhuang, Baoping Cheng, Yao Cheng, Yuntao Jin, Renshuai Liu, Chengyang Li, Xuan Cheng,~\IEEEmembership{Member,~IEEE}, Jing Liao,~\IEEEmembership{Member,~IEEE}, Juncong Lin,~\IEEEmembership{Senior Member,~IEEE}
\thanks{
This work was supported in part by the Natural Science Foundation of Fujian Province of China under Grant 2023J05001, in part by the Natural Science Foundation of Xiamen, China under Grant 3502Z20227012, in part by the NSFC under Grant 62077039, in part by the Fundamental Research Funds for the Central Universities under Grant 20720230106, and in part by the GRF grant (CityU 11216122) from the Research Grants Council (RGC) of Hong Kong. \emph{Corresponding author: Xuan Cheng}.

Yixiang Zhuang, Yuntao Jin, Renshuai Liu, Chengyang Li, Juncong Lin and Xuan Cheng are with the School of Informatics, Xiamen University, Xiamen 361005, China (e-mail: jclin@xmu.edu.cn; chengxuan@xmu.edu.cn).

Baoping Cheng and Yao Cheng are with the China Mobile (Hangzhou) Information Technology Co., Ltd., Hangzhou 311121, China.
Jing Liao is with the Department of Computer Science, City University of Hong Kong, Hong Kong 999077, China (e-mail: jingliao@cityu.edu.hk).
}}

\markboth{Journal of \LaTeX\ Class Files,~Vol.~14, No.~8, August~2021}%
{Shell \MakeLowercase{\textit{et al.}}: A Sample Article Using IEEEtran.cls for IEEE Journals}


\twocolumn[{
\renewcommand\twocolumn[1][]{#1}
\maketitle
\begin{center}
    \includegraphics[width=0.96\linewidth]{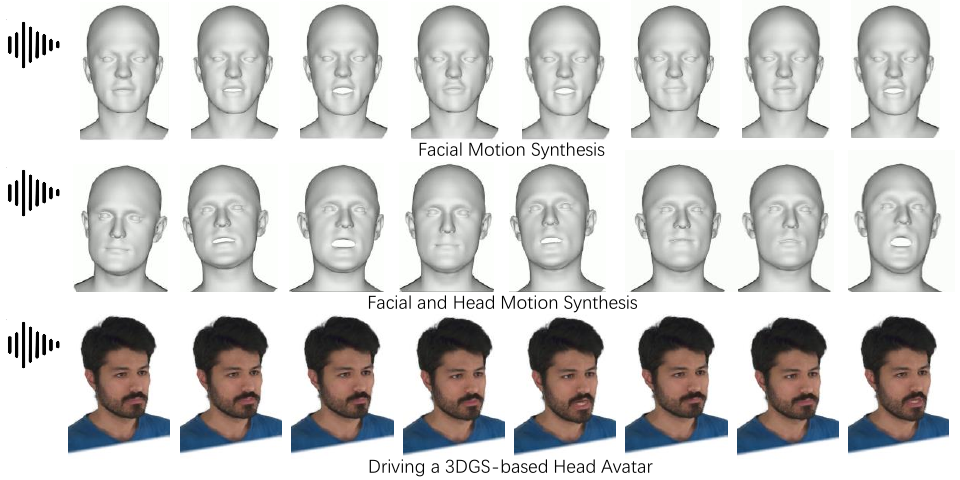}
    \captionof{figure}{Given a speech audio, our method can synthesize facial motion, head motion and drive a 3D Gaussian Splatting \cite{3DGSTOG2023} based head avatar.}
    \label{fig:teaser}
\end{center}
}]

\begin{abstract}
Speech-driven facial animation methods usually contain two main classes, 3D and 2D talking face, both of which attract considerable research attention in recent years. However, to the best of our knowledge, the research on 3D talking face does not go deeper as 2D talking face, in the aspect of lip-synchronization (lip-sync) and speech perception. To mind the gap between the two sub-fields, we propose a learning framework named Learn2Talk, which can construct a better 3D talking face network by exploiting two expertise points from the field of 2D talking face. Firstly, inspired by the audio-video sync network, a 3D sync-lip expert model is devised for the pursuit of lip-sync between audio and 3D facial motion. Secondly, a teacher model selected from 2D talking face methods is used to guide the training of the audio-to-3D motions regression network to yield more 3D vertex accuracy. Extensive experiments show the advantages of the proposed framework in terms of lip-sync, vertex accuracy and speech perception, compared with state-of-the-arts. Finally, we show two applications of the proposed framework: audio-visual speech recognition and speech-driven 3D Gaussian Splatting based avatar animation. The project page of this paper is: \href{https://lkjkjoiuiu.github.io/Learn2Talk/}{https://lkjkjoiuiu.github.io/Learn2Talk/}.
\end{abstract}


\section{Introduction}
Human speech is by nature bimodal \cite{survey2022}, visual and audio. The technology of generating realistic 2D/3D facial animations from audio can serve as a bridge between digital realm and human emotions, enabling a more natural and intuitive form of communication. In recent years, significant progress has been witnessed in the field of speech-driven facial animation, due its widely potential applications across immersive interactions, news reporting and live e-commerce, and the recent boom of generative models, ranging from GANs \cite{GANs2014, StyleGAN2019, EMEF2023} to Diffusion Models \cite{DDPM2020, StableDiffusion, DiffSFSR2024}.


There are mainly two research lines in the filed of speech-driven facial animation, 2D and 3D talking face. The 2D talking face methods \cite{YouSaidThat2017, LipMovements2018, Wave2lip2020, LipGAN2019, FaceAnimationGAN2020, PCAVS2021, ATVGnet2019, Wav2Mov2020, MakeltTalk2020, LiveSpeechPportraits2021, FlowTalkingFace2021, SadTalker2023, DINet2023} usually generate the lip motion or head motion in the pixel space (e.g. image, video) to match the given input audio stream, while the 3D talking face methods use the temporal 3D vertex data (e.g. 3D face template \cite{VOCA2019, MeshTalk2021, FaceFormer2022, CodeTalker2023}, blendshape parameters \cite{Visemenet2018, Voice2Face2022, FaceDiffuser2023}) to represent facial motion. Compared with 2D talking face methods, 3D talking face methods can synthesize more subtle lip movements, since the fine-grained lip shape correction can be better performed in the 3D space. Additionally and importantly, 3D facial animation is perhaps a more appropriate solution in the standardized virtual human production workflow, where the 3D model of virtual human is firstly made by the modeling artists and then be driven by different modalities, e.g. speech-driven \cite{TalkShowCVPR24, libinliuSIG22, LDASig23}, text-driven \cite{TEACH22, UDECVPR23, T2MCVPR23} and video-driven \cite{MAEDICCV21, SimPoECVPR21, MPSNetCVPR22}. On the other hand, however, the research on 2D talking face \cite{SadTalker2023, Wave2lip2020} is going deeper in the aspect of lip-sync and speech perception, both of which are important to speech-driven facial animation.

There are evidences in the research community to support our claims. 
1) The state-of-the-art 3D talking face methods \cite{FaceFormer2022, CodeTalker2023} usually use the 3D reconstruction errors in all lip vertices (taking the maximum) to measure the lip-sync, whereas the 2D talking face methods prefer to use the pre-trained SyncNet \cite{SyncNet2016} to estimate the time offset between the audio and the generated video. SyncNet \cite{SyncNet2016} is a direct measurement of the temporal relationship between speech audio and facial motion, while the 3D reconstruction error well represents per-frame 3D accuracy of lip vertices.
2) Through the visual comparisons between the talking face videos rendered by 3D methods like FaceFormer \cite{FaceFormer2022} and CodeTalker \cite{CodeTalker2023}, and produced by 2D methods like SadTalker \cite{SadTalker2023}, we find that the 2D talking face videos usually show more expressive movements of the mouth corresponding to the speech. For example, a speech-aware talking face should display obvious lip-roundness when speaking a rounded-vowel, such as /o/ /u/.

\begin{figure}
    \centering
    \includegraphics[width=0.46\textwidth]{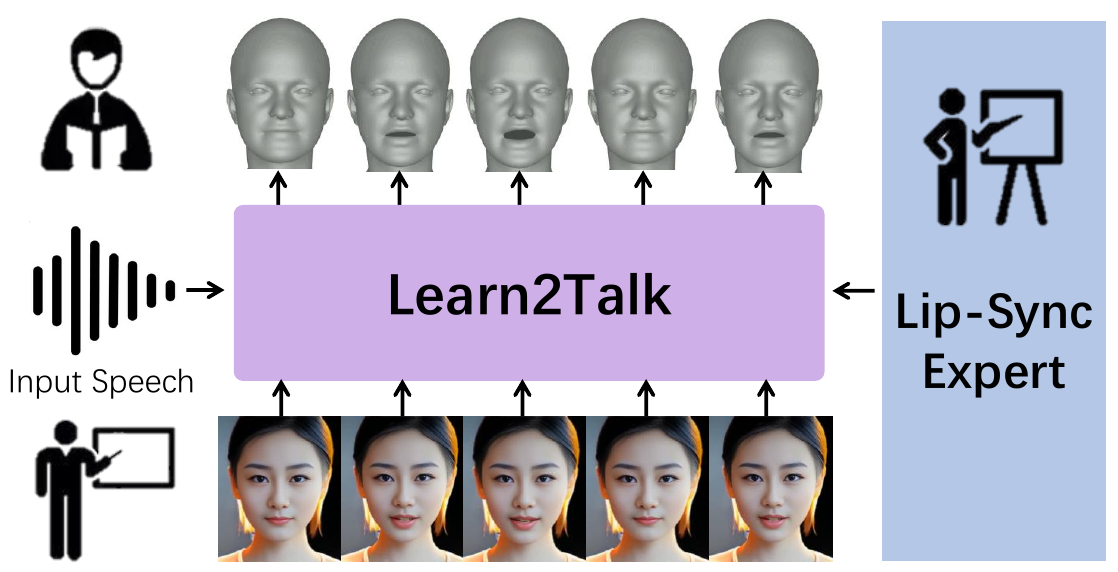}
    \caption{Concept diagram of Learn2Talk framework.}
    \label{fig:banner}
\end{figure}

To mind the gap between the two sub-fields in speech-driven facial animation, we propose a novel framework named \emph{Learn2Talk}, which can construct a better audio-to-3D motion regression network by learning the expertise from the field of 2D talking face. To realize the framework, we propose two innovative network designs in the training of the audio-to-3D motion regression network. 
1) Inspired by SyncNet \cite{SyncNet2016} which is widely used in 2D talking face methods, we develop a 3D lip-sync expert model named \emph{SyncNet3D}, to impose the lip-syncing constraint in training and evaluate the sync between audio and regressed 3D motions in testing.
2) A pre-trained 2D talking face network is employed as the teacher model to supervise the student model, namely the regression network, through the \emph{lipreading constraint} implemented by the pre-trained lipreading network \cite{LSRS3_2022} and differentiable renderer. As shown in Fig. \ref{fig:teaser}, our proposed framework achieves the synthesis of facial and head motion, and the driving of head avatar. The concept diagram of the framework is shown in Fig. \ref{fig:banner}.

Finally, the contributions of this paper are summarized as follows:
\begin{itemize}
    \item \emph{\textbf{Framework.}} We propose a novel learning framework for 3D talking face, which learns the lip-syncing and speech perception capabilities from 2D talking face methods. To the best of our knowledge, this research line has never been explored by previous methods.

    \item \emph{\textbf{Lip-sync.}} We successfully extend SyncNet from the pixels domain to the 3D motions domain. The proposed SyncNet3D can be used as discriminator in training to enhance lip-sync, and as metrics in testing to complement the evaluation of synthesized 3D motions’ quality. This part of work addresses a critical issue existed in almost all 3D motion synthesis methods: \emph{how to explicitly model the sync between 3D motions and input signals?} 
    
    \item \emph{\textbf{Speech Perception.}} We propose to distill the knowledge from 2D talking face methods to the audio-to-3D motion regression model by the lipreading constraint. As such, the predicted 3D facial motions yield more accuracy in lip vertices, thus eliciting similar perception with the corresponding audios. Before the proposing of our method, the lipreading constraint is mainly used in video-driven 3D animation, but \emph{rarely in audio-driven 3D animation.} 
    

    \item  \emph{\textbf{SOTA.}} Both the quantitative comparisons in public datasets, and the extensive visual comparisons on nearly 20 audio clips in the supplementary video, clearly show that our method advances the state of the art.

    \item \emph{\textbf{{Application.}}}
    As an application, our method can be used to drive a full head avatar built by 3D Gaussian Splatting (3DGS) \cite{3DGSTOG2023}, thus enabling the first speech-driven 3DGS-based avatar animation in the research community.
\end{itemize}

\section{Related Work}
\subsection{Speech-driven 2D Facial Animation}
The methods in this field can animate a portrait image or edit a portrait video to match the input driving audio. From the methodological perspective, the methods are roughly categorized into two classes \cite{survey2022}: one-stage framework and two-stage framework. 

The one-stage framework \cite{YouSaidThat2017, LipMovements2018, Wave2lip2020, LipGAN2019, FaceAnimationGAN2020, PCAVS2021} usually generates talking face video from the driving source (e.g. audio-image or audio-video) in an end-to-end fashion. Speech2Vid \cite{YouSaidThat2017} is among the first to explore one-stage framework, which uses two encoders to respectively extract identity features and spoken features from the reference image and driving audio, and then fuses the two features as input into an image decoder to output synthesized video. Besides lip motion, PC-AVS \cite{PCAVS2021} can also generate head motion by using the third encoder to extract the identity-ignore head pose features from another pose source video. Other works put the efforts mostly on developing powerful discriminators to check generation quality in the generative adversarial learning, such as using optical flow stream \cite{LipMovements2018}, synced/unsynced contrastive loss \cite{LipGAN2019}, pre-trained lip-sync expert \cite{Wave2lip2020} and multiple discriminators focused on lip-sync, frame quality and temporality \cite{FaceAnimationGAN2020}. 

Recent advances mostly adopt the two-stage framework \cite{ATVGnet2019, Wav2Mov2020, MakeltTalk2020, LiveSpeechPportraits2021, FlowTalkingFace2021, SadTalker2023}, which contains two cascaded modules: firstly mapping the driving source to intermediate facial parameters by deep neural networks, and then rendering the output video based on the the learned facial parameters. Facial landmarks are the commonly used intermediate representation, since they can capture the facial deformations caused by talking, expressions and head movements. For example, ATVGnet \cite{ATVGnet2019} and Wav2Mov \cite{Wav2Mov2020} use facial landmark PCA components to decrease the effect of none-audio-correlated factors, while MakeItTalk \cite{MakeltTalk2020} uses the relative landmark displacements to generalize to large variety of faces.
As facial landmarks are still in a coupled space, there are also methods that explore the use of 3D facial landmarks \cite{LiveSpeechPportraits2021} and 3D Morphable Model \cite{FlowTalkingFace2021, SadTalker2023}. 
In the aspect of rendering output video, GAN is mostly used \cite{MakeltTalk2020, LiveSpeechPportraits2021, Wav2Mov2020, ATVGnet2019} and is separately trained with previous mapping network. Instead of generating pixels by multiple up-sampling layers, DINet \cite{DINet2023} spatially deform the reference image to match the driving audio, thus generating high-fidelity video. Similarly, SadTalker \cite{SadTalker2023} produces the warping fields from unsupervised 3D keypoints, and then warps the reference image under the guidance of warping fields to generate final video.

\subsection{Speech-driven 3D Facial Animation}
The methods in this field can animate a 3D face model by predicting vertex positions from driving audio. To the best of our knowledge, the amount of research works in this filed is relatively less than speech-driven 2D facial animation. Karras et al. \cite{karras2017audio} take advantage of LSTM to learn the mapping from audio to 3D vertex coordinates and utilize an extra emotion to control the emotional state of the face puppet. VOCA \cite{VOCA2019} extents the model to multiple subjects by using identity conditioning, addressing the issue of specialization for a particular speaker \cite{karras2017audio}. In VOCA, the extracted audio features are concatenated with the onehot vector of a speaker, and then the fused features are decoded to output 3D vertex displacements instead of vertex coordinates. Noting that VOCA exhibits uncanny or static upper-face animation, MeshTalk \cite{MeshTalk2021} proposes a categorical latent space to disentangle audio-correlated and audio-uncorrelated information, so that the plausible motion of uncorrelated facial regions can be synthesized. GDPnet \cite{GDPnet2022} improves on VOCA with more geometry-related deformation, which is achieved by adding the non-linear face reconstruction representation as the guidance of latent space. To tackle the inaccurate lip movement associated with phoneme-level features from short audio windows, FaceFormer \cite{FaceFormer2022} proposes a transformer-based autoregressive model to encode long-term audio contexts, and gains important performance promotion. Aiming at reducing the cross-modal ambiguity, CodeTalker \cite{CodeTalker2023} uses VQ-VAE \cite{VQVAE2017} to firstly learn a codebook by self-reconstruction over real facial motions, and then deploys FaceFormer in the learned discrete motion space. Inspired by the recent progress in Diffusion Models, there are also methods \cite{FaceDiffuser2023, Diffposetalk2023} that adopt the diffusion model to capture the many-to-many relationship between speech and facial motion.

\subsection{Video-driven 3D Facial Animation}
We review the video-driven 3D facial animation methods briefly. The single image 3D face reconstruction methods, such as RingNet \cite{RingNet2019}, Deep3DFace \cite{Deep3DFace2019}, DECA \cite{DECA2021}, EMOCA \cite{EMOCA2022} and DCT \cite{DCT2023}, can be used to generate 3D facial motions through a frame-by-frame fashion on videos. There are also methods \cite{DDE2014, HighFidelityDDE2015, 3Deyegaze2016, 3DFaceRigs2016, LipreadVideo2022} that exploit the dynamic information of monocular face videos to constrain the subject’s facial shape or impose temporal coherence on the face reconstruction. For example, DDE \cite{DDE2014} imposes the temporal regularization on the 3DMM \cite{blanz1999morphable, multilinear3dmm} fitting of per-frame facial landmarks. Later on, Cao et al. \cite{HighFidelityDDE2015} improve on DDE by training a local detail regressor to synthesize the medium frequency facial details in each frame. Built upon the facial landmarks fitting framework, Wang et al. \cite{3Deyegaze2016} and Wang et al. \cite{3DEyeGaze2021} add the tracking of 3D eye gaze by exploring the classification of iris and pupil pixels. Garrido et al. \cite{3DFaceRigs2016} propose a parametric shape prior that encodes the plausible subspace of facial identity and expression variations to reconstruct a fully personalized 3D face rig from input video. The most recent method \cite{LipreadVideo2022} use DECA \cite{DECA2021} to estimate per-frame 3D face model, and then use the lipreading network \cite{LSRS3_2022} to enhance the visual speech-aware perceptual in the generated 3D facial motions.

\begin{figure*}
    \centering
    \includegraphics[width=0.99\textwidth]{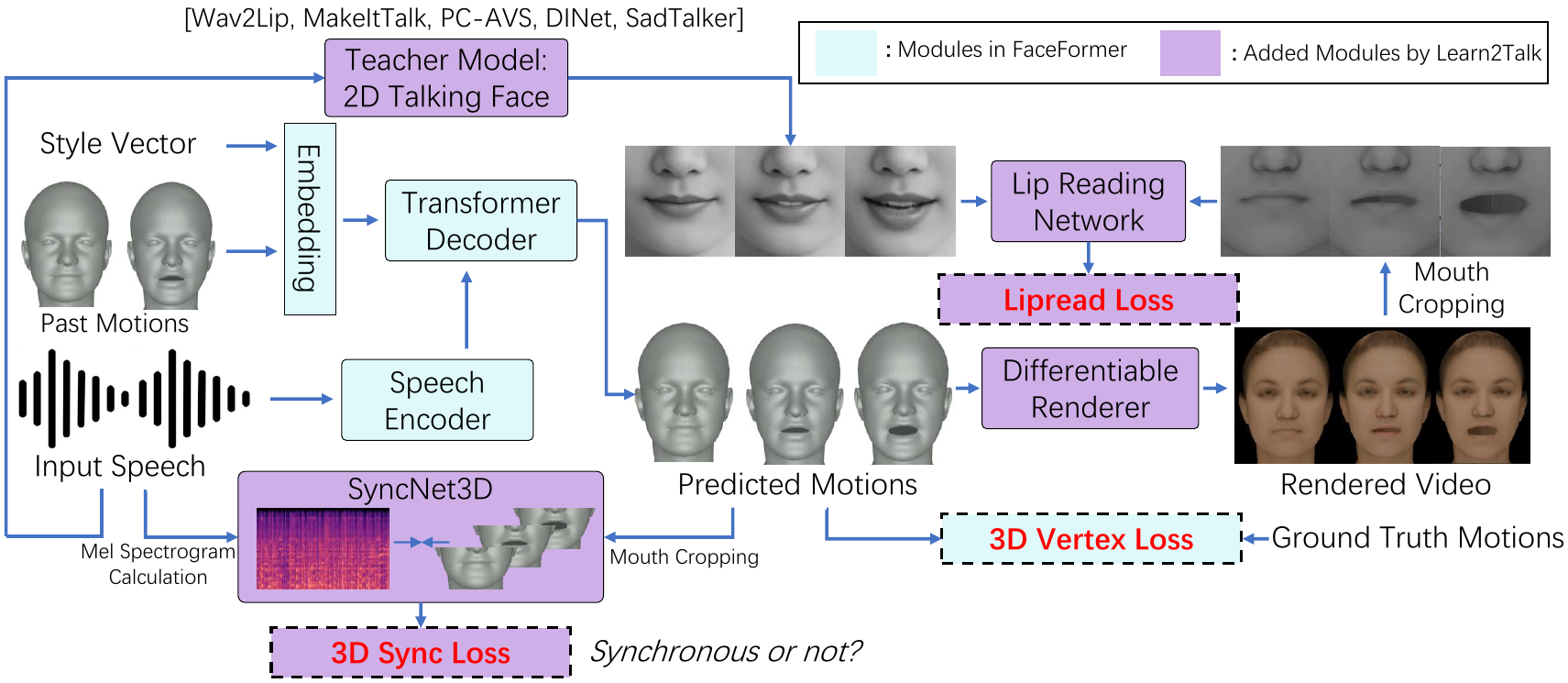}
    \caption{The pipeline of the proposed Learn2Talk framework. In training, all modules are used. In inference, only the student model is used,  including embedding layer, transformed decoder and speech encoder.}
    \label{fig:pipeline}
\end{figure*}

\section{The Proposed Framework}
\subsection{Overview}
\textbf{Problem Setting.} The speech-driven 3D facial animation is formulated as the seq2seq learning problem, which predicts the 3D facial motions from a speech audio. Let $\mathbf{Y}_{1:T} = (\mathbf{y}_1, ..., \mathbf{y}_T)$ be a sequence of 3D facial motions, where $\mathbf{y}_t \in \mathbb{R}^{V\times 3}$ denotes the 3D mesh model of $V$ vertices in frame $t$. Let $\mathbf{X}_{1:T} = (\mathbf{x}_1, ..., \mathbf{x}_T)$ be a sequence of audio snippets, where $\mathbf{x}_t \in \mathbb{R}^{d}$ denotes the snippet of $d$ samples aligned with the corresponding motion frame $\mathbf{y}_t$. The goal is to train a regression network which can sequentially generate $\mathbf{Y}_{1:T}$ from the inputted $\mathbf{X}_{1:T}$, so that the 3D face mesh model can be animated as a lip-synchronized 3D talking face in the inference stage.

\textbf{Pipeline.} Fig. \ref{fig:pipeline} shows the pipeline of the proposed framework. FaceFormer \cite{FaceFormer2022} is chosen as the student model, which predicts 3D facial motions from both audio context and past motions. Meanwhile, a pre-trained 2D talking face network is employed as the teacher model, e.g. Wav2Lip \cite{Wave2lip2020}, SadTalker \cite{SadTalker2023} etc.. The speech audio $\mathbf{X}_{1:T}$ is firstly input to teacher model to produce the image sequence of talking lips. In the other stream, the predicted sequence $\hat{\mathbf{Y}}_{1:T}$ by the student model is input into the differentiable renderer to produce the rendered talking lips. These two streams are contrasted in the pre-trained lipreading network \cite{LSRS3_2022}, thus forming the lipread loss. Besides, the pre-trained SyncNet3D serves as the lip-sync discriminator, where a predicted motion clip $\hat{\mathbf{Y}}_{1:W}$ and an audio clip $\mathbf{X}_{1:W}$ ($W$ denotes window size) are contrasted there to form the 3D sync loss. Together with 3D vertex reconstruction loss, the student model is trained in a supervised manner.

\subsection{Student Model} 
The student model contains two main modules, speech encoder and cross-modal decoder. The speech encoder $E_{\text{speech}}$ adopts the architecture of the state-of-the-art self-supervised pretrained speech model, wav2vec 2.0 \cite{wav2vec2020}, which consists of an audio feature extractor and a multi-layer transformer encoder \cite{Attention2017}. The audio feature extractor is a temporal convolutions network (TCN), which converts the speech of raw waveform into feature vectors. The transformer encoder is a stack of multi-head self-attention and feed-forward layers, which converts the audio feature vectors into contextualized speech representations. 

The cross-modal decoder $D_{\text{cross-modal}}$ contains an embedding block and a multi-layer transformer decoder equippe with both self-attention and cross-attention. The embedding block combines the past facial motions $\hat{\mathbf{Y}}_{1:t-1}$ and the style embedding via:
\begin{equation}
    \mathbf{F}_{emb}^{1:t-1} = \mathcal{P}_{\theta}(\hat{\mathbf{Y}}_{1:t-1}) + \mathbf{B} \cdot \frac{\mathbf{s}}{||\mathbf{s}||_1},
\end{equation}
where $\mathcal{P}_{\theta}$ denotes a linear projection layer, and $\mathbf{B}$ denotes the learnable basis vector that span the style space linearly. The transformer decoder is equipped with the causal self-attention to learn the dependencies between each frame in the context of the past facial motions, and the cross-modal attention to align the audio and motion modalities. The newly predicted motion $\hat{\mathbf{y}}_t$ is used to update the past motions as $\hat{\mathbf{Y}}_{1:t}$, in preparation for the next prediction. The recursive process can be formally written as:
\begin{equation}
    \hat{\mathbf{y}}_t = D_{\text{cross-modal}}(E_{\text{speech}}(\mathbf{X}_{1:T}), \mathbf{s}, \hat{\mathbf{Y}}_{1:t-1}).
\end{equation}

In the training, the TCN and transformer encoder in $E_{\text{speech}}$ are initialized with the pre-trained wav2vec 2.0 weights to benefit from the speech representation learning from large-scale corpora. The parameters in TCN are fixed, whereas transformer encoder in $E_{\text{speech}}$, the embedding block and transformer decoder in $D_{\text{cross-modal}}$ are learnable. The training loss is the Mean Squared Error (MSE) between the outputs $\hat{\mathbf{Y}}_{1:T} = (\hat{\mathbf{y}}_1, ..., \hat{\mathbf{y}}_T)$ and the ground truth $\mathbf{Y}_{1:T} = (\mathbf{y}_1, ..., \mathbf{y}_T)$: 
\begin{equation}
\label{eq:vertexloss}
    \mathcal{L}_{\text{vertex}} = \sum_{t=1}^T\sum_{v=1}^V||\hat{\mathbf{y}}_{t,v}-\mathbf{y}_{t,v}||_2.
\end{equation}
This loss is computed when all frames in the 3D facial motion sequence are produced.

\begin{table}[]
\centering
\begin{tabular}{l|ll|lll}
\hline
\multicolumn{1}{c|}{} & \multicolumn{2}{c|}{Lip-sync}  & \multicolumn{3}{c}{Video Quality}
\\ 
\cline{2-6} 
{\color[HTML]{000000} }                      & \multicolumn{1}{c}{{LSE-D$\downarrow$}} & \multicolumn{1}{c|}{{LSE-C$\uparrow$}}  & \multicolumn{1}{c}{{FID$\downarrow$}} & \multicolumn{1}{c}{{CPBD$\uparrow$}} & \multicolumn{1}{c}{{CSIM$\uparrow$}} 
\\ 
\hline
{\color[HTML]{000000} Real Video}               & {\color[HTML]{000000} 6.982}                  & {\color[HTML]{000000} 8.211}             & {\color[HTML]{000000} 0.0000}                  & {\color[HTML]{000000} 0.428}                  & {\color[HTML]{000000} 1.000}                     \\
{\color[HTML]{000000} Wave2Lip$\dag$}           & {\textbf{5.535}}                  & {\textbf{10.211}}                   
& {\color[HTML]{000000} 21.725}                  & {\color[HTML]{000000} 0.368}                  & {\color[HTML]{000000} 0.849}                     
\\
{\color[HTML]{000000} PC-AVS}              & {\color[HTML]{000000} 6.355}                  & {\color[HTML]{000000} 9.053}                & {\color[HTML]{000000} 69.127}                  & {\color[HTML]{000000} 0.206}                  & {\color[HTML]{000000} 0.683}                     \\
{\color[HTML]{000000} MakeItTalk}              & {\color[HTML]{000000} 10.059}                  & {\color[HTML]{000000} 5.110}                  & {\color[HTML]{000000} 28.243}                  & {\color[HTML]{000000} 0.283}                  & {\color[HTML]{000000} 0.838}                     \\
{\color[HTML]{000000} AVCT}                   & {\color[HTML]{000000} 10.055}                  & {\color[HTML]{000000} 4.932}         & {\color[HTML]{000000} 22.432}                  & {\color[HTML]{000000} 0.295}                  & {\color[HTML]{000000} 0.811}                     \\
{\color[HTML]{000000} SadTalker}                  & {\color[HTML]{000000} 7.772}                  & {\color[HTML]{000000} 7.290}                & {\textbf{22.057}}                  & {\textbf{0.335}}                  & {\textbf{0.843}}                     
\\
\hline
{\color[HTML]{000000} Real Video*}                  & {\color[HTML]{000000} 6.499}                  & {\color[HTML]{000000} 8.993}               & {\color[HTML]{000000} N/A}                  & {\color[HTML]{000000} N/A}                  & {\color[HTML]{000000} N/A}    \\
{\color[HTML]{000000} ATCG*}                  & {\color[HTML]{000000} 8.422}                  & {\color[HTML]{000000} 7.061}           & {\color[HTML]{000000} N/A}                  & {\color[HTML]{000000} N/A}                  & {\color[HTML]{000000} N/A}     \\
{\color[HTML]{000000} DINet*}                  & {\color[HTML]{000000} 6.842}                  & {\color[HTML]{000000} 8.377}           & {\color[HTML]{000000} N/A}                  & {\color[HTML]{000000} N/A}                  & {\color[HTML]{000000} N/A}     \\
\hline
\end{tabular}
\caption{The quantitative evaluation results of seven 2D talking face methods over the HDTF dataset \cite{FlowTalkingFace2021}. The best score in each column is marked with \textbf{bold}.} 
\label{tab:2DTalkingFace}
\end{table}

\subsection{Teacher Models Selection}
There are many alternative state-of-the-art 2D talking face methods, but we only need the one-shot, head motion excluded methods to make them compatible with our framework. Tab. \ref{tab:2DTalkingFace} shows the evaluation results of seven 2D talking face methods over the HDTF dataset \cite{FlowTalkingFace2021}, including Wav2Lip \cite{Wave2lip2020}, PC-AVS \cite{PCAVS2021}, MakeItTalk \cite{MakeltTalk2020}, AVCT \cite{AVCT2022}, SadTalker \cite{SadTalker2023}, ATCG \cite{ATVGnet2019} and DINet \cite{DINet2023}. HDTF contains in-the-wild videos collected 720P or 1080P resolution. Two classes of metrics are used to quantitatively evaluate the lip-sync and video quality of the generated talking face videos, including distance score (LSE-D)  \cite{SyncNet2016}, confidence score (LSE-C) \cite{SyncNet2016}, Frechet Inception Distance (FID) \cite{FID2017}, cumulative probability blur detection (CPBD) \cite{CPBD2011} and cosine similarity of identity (CSIM). The statistics come from 
the experiment conducted in \cite{SadTalker2023} and another similar experiment in \cite{DINet2023} (denoted by $*$). Since the experiments in \cite{SadTalker2023, DINet2023} selected different videos as testing data, there is a slight bias in the metrics LSE-D and LSE-C. Although Wav2Lip$\dag$ acheives the best quantitative results in video quality, it only animates the lip region while other regions are the same as the original frame. Hence, actually, it's SadTalker that acheives the best video quality.

From the statistics in Tab. \ref{tab:2DTalkingFace}, we can observe that Wave2Lip performs well in the lip-sync, while SadTalker exhibits good performance in high quality video generation. Besides lip-sync, video quality should be considered in selecting teacher models, since the quality of generated video has much impact on the transferring process that is based on the lip reading from image sequence. Hence, we select five methods as the teacher model $G_{\text{teach}}$, including Wav2Lip, SadTalker, DINet, PC-AVS and MakeItTalk, based on their overall performance in lip-sync and video quality.

\subsection{SyncNet3D}
Current state-of-the-art 3D talking face methods \cite{VOCA2019, MeshTalk2021, GDPnet2022, FaceFormer2022, CodeTalker2023, FaceDiffuser2023, Diffposetalk2023} usually use the 3D vertex reconstruction loss defined in Eq. \ref{eq:vertexloss} as the main training objective. In the reference stage, the 3D lip vertex error (LVE), which calculates the maximal $L_2$ distance of all lip vertices to the ground truth in each frame and then takes average over all frames, is used as the main metric to measure lip-sync. The relationship in granular level between audio and 3D facial motion should be explicitly modeled, e.g. the audio lags the motion by 60ms or the audio leads the motion by 25ms, which can further support supervising the network and measuring the lip-sync in results.

Based on the above observations, we propose the SyncNet3D which is accurate in detecting sync error in 3D facial motions and its corresponding audio. The inputs to SyncNet3D contain a mesh stream and a audio stream. In the mesh stream, a window of $W$ consecutive frames of    3D mouth mesh model, $\mathbf{M}_{1:W} = (\mathbf{m}_1, ..., \mathbf{m}_W)$, is fetched from $\mathbf{Y}_{1:T}$ by a sliding window, and then be encoded by a transformer and three-layer MLP into the mesh embedding $\mathbf{M}_{emd}^{1:W}$. In the audio stream, a window of $W$ frames of audio snippets, $\mathbf{X}_{1:W} = (\mathbf{x}_1, ..., \mathbf{x}_W)$, are fetched from $\mathbf{X}_{1:T}$ by the sliding window, and then be converted to Mel spectrogram. The Mel spectrogram is futher encoded by five convolutional layers and two-layer MLP into the audio embedding $\mathbf{X}_{emd}^{1:W}$, which has the same shape with $\mathbf{M}_{emd}^{1:W}$. The contrastive loss \cite{SyncNet2016} between $\mathbf{M}_{emd}^{1:W}$ and $\mathbf{X}_{emd}^{1:W}$ is used as the training objective, aiming at making $\mathbf{M}_{emd}^{1:W}$ and $\mathbf{X}_{emd}^{1:W}$ similar for genuine pairs, and different for false pairs. The contrastive loss is defined as:
\begin{equation}
\label{eq:contrastive1}
\begin{split}
    \mathcal{L}_{\text{con1}} = \sum_{n=1}^N(y)d^2 + (1-y) \text{max}(p-d, 0)^2, \\
    d = ||\mathbf{M}_{emd}^{1:W} - \mathbf{X}_{emd}^{1:W}||_2, \quad\quad\quad\quad
\end{split}    
\end{equation}
where $N$ denotes the number of training samples, $y\in[0,1]$ denotes the binary similarity metric and $p$ denotes the margin parameter.

We pre-train a SyncNet3D using Eq. \ref{eq:contrastive1} to measure the lip-sync in the quantitative evaluation (Sect. \ref{sect:datasets_evaluation}), and pre-train another SyncNet3D which serves as the parameter-fixed discriminator in the training of Learn2Talk (Sect. \ref{sect:TrainingLosses}), using a different contrastive loss \cite{Wave2lip2020}:
\begin{equation}
\label{eq:contrastive2}
\begin{split}
    \mathcal{L}_{\text{con2}} =- \frac{1}{N} \sum_{n=1}^N(y)\log_{}{d}  + (1-y) log_{}{(1-d)}, \\
    d = \text{CosSim}(\mathbf{M}_{emd}^{1:W}, \mathbf{X}_{emd}^{1:W}),
    \quad\quad\quad\quad
\end{split}    
\end{equation}
where $\text{CosSim}(\cdot, \cdot)$ denotes the cosine similarity between two vectors. The cosine similarity will be normalized to the range of $[0, 1]$, thus indicating the probability of being synchronous. The reason of pre-training another SyncNet3D using Eq. \ref{eq:contrastive2} is that, the optimization is easier to converge with the cosine similarity than the $L_2$ distance. In contrast, it's easier to separate the features of genuine and false pairs with the $L_2$ distance and the maximum operation in Eq. \ref{eq:contrastive1}.

\subsection{Training Losses}
\label{sect:TrainingLosses}
\textbf{Lipread Loss.} The teacher model supervises the training of Learn2Talk through the lipread loss. We use the lipreading network \cite{LSRS3_2022} pre-trained on the Lip Reading in the Wild 3 (LRS3) dataset \cite{LSRS3_2022} to compute the lipread loss. The network takes a sequence of grayscale images cropped around the mouth as input, and outputs a predicted sequence of characters. The model architecture consists of a 3D convolutional kernel, a 2D ResNet-18, a 12-layer conformer \cite{Conformer2020}, and lastly a transformer decoder layer that outputs the predicted sequence. The network is trained by using Connectionist Temporal Classification equipped with attention. When used as the encoder $E_{\text{lipread}}$ for talking face, the network discards the 12-layer conformer and the transformer decoder layer, and output the embedding from the 2D ResNet-18 layer. 

\begin{figure}
    \centering
    \includegraphics[width=0.46\textwidth]{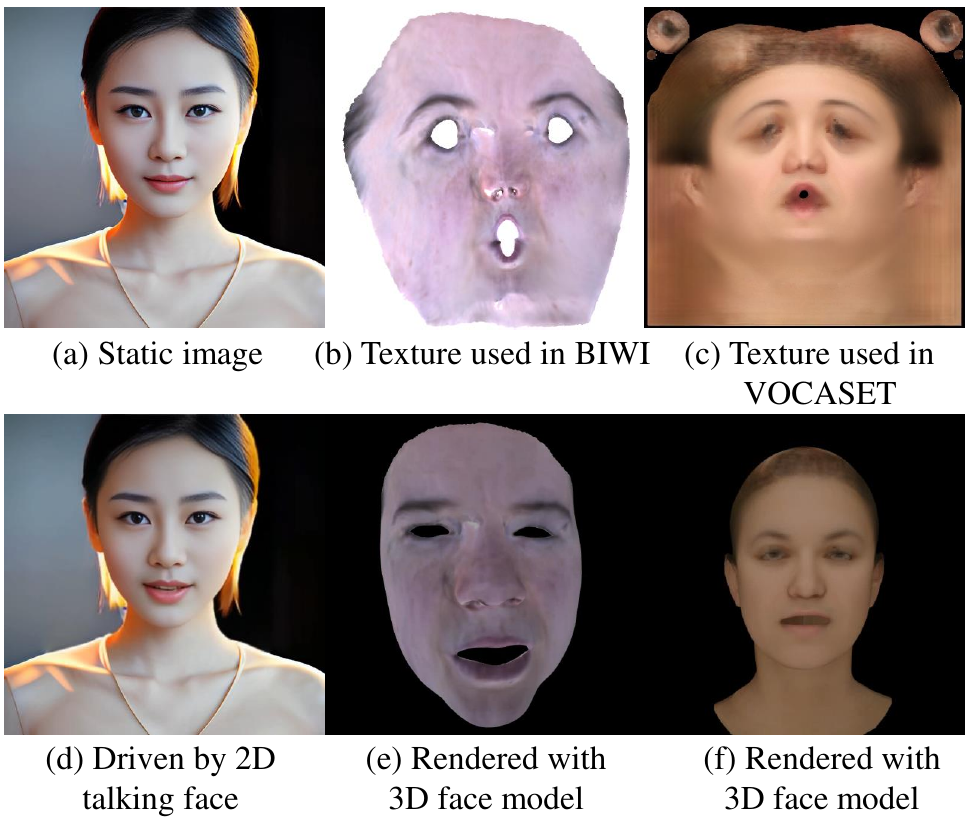}
    \caption{The static image (a) is used in one-shot 2D talking face methods to generate animated face (d). The facial textures (b)(c) are used in rendering 3D facial motions to videos (e)(f).}
    \label{fig:textures}
\end{figure}

The speech audio $\mathbf{X}_{1:T}$ is input to the teacher model $G_{\text{teach}}$ to produce the image sequence of 2D talking face. Then, the grayscale images are cropped around mouth and input to $E_{\text{lipread}}$ to get the embedding $E_{\text{lipread}}(G_{\text{teach}}(\mathbf{X}_{1:T}))$. Meanwhile, the predicted 3D facial motions $\hat{\mathbf{Y}}_{1:T}$ together with a facial texture shown in Fig. \ref{fig:textures}, are injected into the differentiable renderer $DR$ to produce the rendered image sequence of 3D talking face. Similarly, the grayscale images are cropped around mouth and input to $E_{\text{lipread}}$ to get the embedding $E_{\text{lipread}}(DR(\hat{\mathbf{Y}}_{1:T}))$. Finally, the lipread loss is defined as the cosine similarity between two embeddings:
\begin{equation}
\label{eq:lipread}
\begin{split}
    \mathcal{L}_{\text{lipread}} = \text{CosSim}(E_{\text{lipread}}(G_{\text{teach}}(\mathbf{X}_{1:T})), \\ 
    E_{\text{lipread}}(DR(\hat{\mathbf{Y}}_{1:T}))).
\end{split}    
\end{equation}
This loss is computed when all frames $\hat{\mathbf{Y}}_{1:T}$ in the 3D facial motion sequence have been produced. Thanks to the differentiable renderer, the gradients in $\mathcal{L}_{\text{lipread}}$ can be back-propagated in the optimization.

\textbf{3D Sync Loss.} The SyncNet3D pre-trained by Eq. \ref{eq:contrastive2} is used as the encoder to get two embeddings $\hat{\mathbf{M}}_{emd}^{t+1:t+W}, \mathbf{X}_{emd}^{t+1:t+W}$ respectively from $\hat{\mathbf{Y}}_{t+1, t+W}$, $\mathbf{X}_{t+1, t+W}$. Then, the 3D sync loss is defined as:
\begin{equation}
\label{eq:syncloss}
\begin{split}
    \mathcal{L}_{\text{sync}} = \sum_{t=0}^{T-W}\text{CosSim}(\hat{\mathbf{M}}_{emd}^{t+1:t+W}, \mathbf{X}_{emd}^{t+1:t+W}).
\end{split}    
\end{equation}

\textbf{Overall Loss.} The overall loss is defined as:
\begin{equation}
\label{eq:overall}
\begin{split}
    \mathcal{L} = \mathcal{L}_{\text{vertex}} + \lambda_1\mathcal{L}_{\text{sync}} + \lambda_2\mathcal{L}_{\text{lipread}}, 
\end{split}    
\end{equation}
where $\lambda_1, \lambda_2$ are two hyper-parameters.

\subsection{Head Motion Synthesis}
Apart from the facial motion synthesis, the head motion synthesis from audio can results in more vivid 3D facial animation. To achieve this goal, we directly adopt the PoseVAE network proposed in SadTalker \cite{SadTalker2023}, and incorporate it in our framework by applying the predicted rotation and translation vectors on the 3D mesh model $\mathbf{y}_t$ in each frame. PoseVAE uses the VAE model \cite{VAE2024} conditioned by audio clip to learn the distribution of the 6-dimensional head movement. As the head motion synthesis is performed independently and parallelly with the facial motion synthesis, it's a optional process in our framework.

\section{Experiments}
\label{sect:experiments}

\subsection{Datasets and Evaluation Metrics}
\label{sect:datasets_evaluation}
\textbf{Datasets.} We use two widely used datasets in the field of 3D talking face, BIWI \cite{BIWI2010} and VOCASET \cite{VOCA2019}, to train and test different methods in the experiments. Both datasets contain the sequences of 3D face scans together with corresponding utterances spoken in English. 1) \textbf{BIWI} is a 3D audio-visual corpus of affective speech and facial expression in the form of dense dynamic 3D face geometries, containing totally 40 sentences uttered by 14 subjects. Each sentence was recorded twice, with and without emotion, and is 4.67 seconds long averagely. The 3D face dynamics are captured at 25fps, each with 23,370 vertices and registered topology. We only use the emotional subset and follow the data splits in \cite{FaceFormer2022}. Specifically, the training set (BIWI-Train) contains 192 sentences, the validation set (BIWI-Val) contains 24 sentences, the testing set (BIWI-Test-A) contains 24 sentences spoken by 6 seen subjects.
2) \textbf{VOCASET} is comprised of 480 paired audio-visual sequences recorded from 12 subjects. 255 unique sentences partially shared among different speakers, are included in VOCASET. Each sequence is captured at 60 fps and is between 3 and 4 seconds averagely. Each 3D face mesh is registered to the FLAME \cite{FLAME2017} topology with 5,023 vertices. We adopt the same training (VOCA-Train), validation (VOCA-Val), and testing (VOCA-Test) splits as VOCA \cite{VOCA2019} and FaceFormer \cite{FaceFormer2022} for fair comparisons.

\textbf{Evaluation Metrics.} To quantitatively evaluate the different methods in terms of 3D lip-sync and 3D vertex reconstruction quality, we adopt four metrics:
\begin{itemize}
\item[--] LSE-D: the lip-sync error distance calculated by the pre-trained SyncNet3D with Eq. \ref{eq:contrastive1}. A lower LSE-D denotes a higher match of audio and 3D motions.
\item[--] LSE-C: the lip-sync error confidence calculated by the pre-trained SyncNet3D with Eq. \ref{eq:contrastive1}. Higher the confidence, the better correlation of audio and 3D motions.
\item[--] LVE: the lip vertex error proposed in \cite{MeshTalk2021}. It calculates the maximal distance of all 3D lip vertices to the ground truth in each frame and then takes average over all frames. A lower LVE denotes a better 3D lip vertex reconstruction quality.
\item[--] FDD: the upper-face dynamics deviation proposed in \cite{CodeTalker2023}. FDD calculates the deviation of temporal variation between the predicted 3D upper faces and ground truth. The lower FDD, the higher consistency with the trend of upper facial dynamics.
\end{itemize}

LVE and FDD, calculated in the 3D space, can't comprehensively measure the lip-sync, since the speech audios are not involved. As the necessary complements, LSE-D and LSE-C measure the lip-sync in another aspect, where the relationship between speech audio and 3D motions is modeled. Summarily, both types of metrics should be adopted in the quantitative evaluation.

\subsection{Implementations}
\textbf{Learn2Talk.} The student model in Learn2Talk is trained with Adam optimizer. The learning rate is set to $1e-4$. We set $\lambda_1 = 5E-6, \lambda_2 = 1E-2$ in BIWI, and set $\lambda_1 = 5E-4, \lambda_2 = 5E-7$ in VOCASET, as the 3D face models used are different. The training duration is 100 epochs, which takes about $3\sim5$ hours on a NVIDIA RTX 4090 GPU.

\textbf{SyncNet3D.} SyncNet3D is separately trained on BIWI-Train and VOCA-Train. The sliding window size $W$ is set as 5 in BIWI and 6 in VOCASET, with the consideration of 25fps and 30fps in two datasets. It takes about $12\sim15$ hours to train SyncNet3D on a NVIDIA RTX 4090 GPU.

\subsection{Study on SyncNet3D}
\begin{figure}
    \centering
    \includegraphics[width=0.45\textwidth]{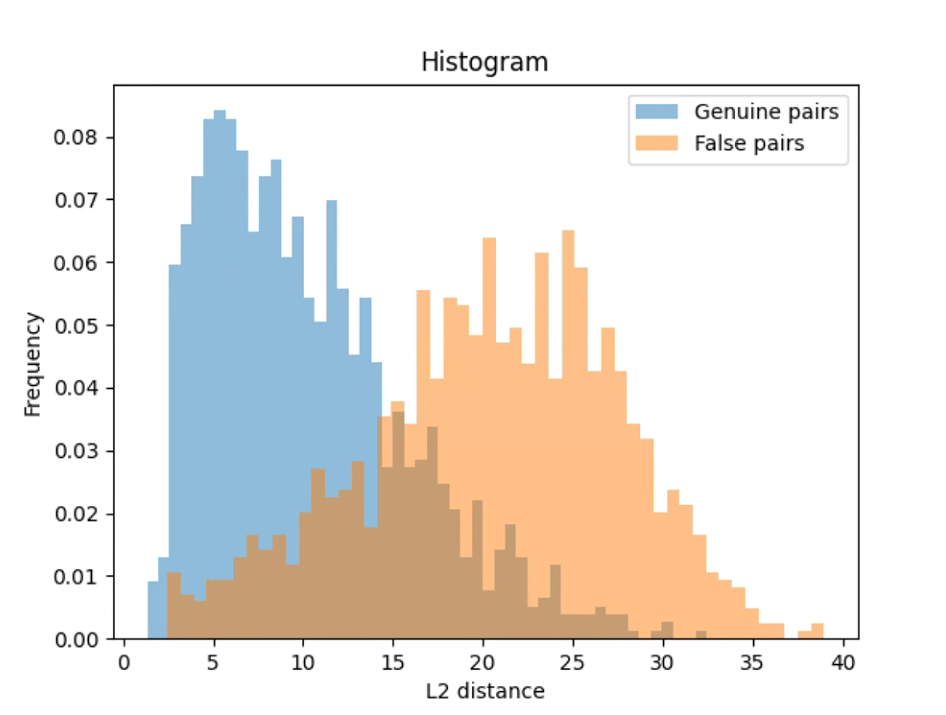}
    \caption{The distribution of $L_2$ distances for genuine and false audio-3D motions pairs.}
    \label{fig:histogram}
\end{figure}

We conduct an experiment to validate the proposed SyncNet3D, in the aspect of determining lip-sync error in audio-3D sequences. Specially, we firstly train a SyncNet3D with Eq. \ref{eq:contrastive1} using BIWI-Train data, in which the genuine audio-3D pairs are generated by taking a $W$-frame audio clip and the corresponding 3D motion clip, the false audio-3D pairs are generated by only shifting the audio randomly by up to 2 seconds.
Then, we construct the test dataset from BIWI-Test-A, using the same strategy. Finally, we use the pre-trained SyncNet3D to determine the lip-sync error in the test dataset.
The $L_2$ distance (Eq. \ref{eq:contrastive1}) between audio embedding and 3D motion embedding is used as the similarity metric. The histogram in Fig. \ref{fig:histogram} shows the distribution of the $L_2$ distances, which indicates that our SyncNet3D can distinguish being synchronous or asynchronous. 

\subsection{Study on Teacher Models Selection}
As five 2D talking face methods are selected as the teacher model, we conduct an experiment on BIWI to show the teaching quality of the five models on student model. We use the public pre-trained networks of Wav2Lip \cite{Wave2lip2020}, PC-AVS \cite{PCAVS2021}, MakeItTalk \cite{MakeltTalk2020}, DINet \cite{DINet2023} and SadTalker \cite{SadTalker2023} to produce five sets of 2D talking videos, which take the audios in BIWI-Train as input. Then, the student model is trained on BIWI-Train, together supervised by the set of 2D talking videos. Note that, in the training, only one set of 2D talking videos is used as supervisor. In order to emphasize the impact of teacher model, we discard the 3D sync loss. If none teacher model is used, our method degrades to FaceFormer. The quantitative evaluation results are reported in Tab. \ref{tab:teacher_selection}. It could be observed that: 1) all teacher models can improve the 3D reconstruction ability of the student model, especially SadTalker; 2) except for PC-AVS, all teacher models have a negative impact on the metrics about lip-sync. The lipread loss that we use to supervise the student model, is once computed on a whole sequence (averagely 4.67s in BIWI) of the predicted 3D motions. The lipreading network can be viewed as a global feature extractor for 3D motions, but LSE-D and LSE-C measure the sync in a clip (0.2s in BIWI) and then average all clips. Hence, the supervision in global level may hinder the lip-sync measured in local level. More analysis about the relationship between lip-sync and lipread is provided in the Ablation Study.

Through this study on teacher selection, we choose SadTalker as the only teacher model for the rest of experiments, considering its overall teaching performance in 3D vertex reconstruction and 3D lip-sync. The ability of lip-sync can be further improved by adding 3D sync loss.

\begin{table}[]
\centering
\begin{tabular}{l|ll|ll}
\hline
\multicolumn{1}{c|}{{\color[HTML]{000000} }} & \multicolumn{2}{c|}{{\color[HTML]{000000} 3D Lip-sync}}  & \multicolumn{2}{c}{{\color[HTML]{000000} 3D Vertex Quality}}                                                                                                 \\ 
\cline{2-5} 
{\color[HTML]{000000} }                      & \multicolumn{1}{c}{{LSE-D$\downarrow$}} & \multicolumn{1}{c|}{{LSE-C$\uparrow$}}  & \multicolumn{1}{c}{{LVE$\downarrow$}} & \multicolumn{1}{c}{{FDD$\downarrow$}} \\ \hline
{\color[HTML]{000000} Learn Wav2Lip}               & {\color[HTML]{000000} 9.265}                  & {\color[HTML]{000000} 8.865}             & {\color[HTML]{000000} 5.2399}                  & {\color[HTML]{000000} 5.2150}                                       
\\
{\color[HTML]{000000} Learn PC-AVS}           & {\color[HTML]{000000} \textbf{9.027}}                  & {\color[HTML]{0000FF}\textbf{9.026}}                   & {\color[HTML]{000000} 5.1226}                  & {\color[HTML]{000000} 4.3717}                   \\
{\color[HTML]{000000} Learn MakeItTalk}              & {\color[HTML]{000000} 9.169}                  & {\color[HTML]{000000} 9.009}                & {\color[HTML]{000000} 4.9816}                  & {\color[HTML]{000000} 4.0992}                  
\\
{\color[HTML]{000000} Learn DINet}              & {\color[HTML]{000000} 9.370}                  & {\color[HTML]{000000} 8.745}                  & {\color[HTML]{0000FF} \textbf{4.9205}}                  & {\color[HTML]{0000FF} \textbf{3.5140}}                 
\\
{\color[HTML]{000000} Learn SadTalker}                   & {\color[HTML]{000000} 9.434}                  & {\color[HTML]{000000} 8.933}         & {\textbf{4.6971} }                  & {\textbf{3.4083}}                  
\\
{\color[HTML]{000000} Learn None (FaceFormer)}                   & {\color[HTML]{0000FF}\textbf{9.101}}                  & {\textbf{9.182}}         & {\color[HTML]{000000} 5.3388}                  & {\color[HTML]{000000} 4.4103}
\\
\hline
\end{tabular}
\caption{Study on Teacher Models Selection. The quantitative evaluation results over BIWI-Test-A dataset are reported. The measuring unit is $10^{-4}$ mm in LVE, and $10^{-5}$ mm in FDD. The best score in each column is marked with \textbf{bold}, and the second best score is marked with {\color[HTML]{0000FF}\textbf{blue}}.} 
\label{tab:teacher_selection}
\end{table}

\begin{table}[]
\centering
\begin{tabular}{l|l|ll|ll}
\hline
\multirow{2}{*}{Datasets} & \multirow{2}{*}{Methods} & \multicolumn{2}{c|}{3D Lip-sync}  & \multicolumn{2}{c}{3D Vertex Quality}\\ 
\cline{3-6} 
{} & {} & \multicolumn{1}{c}{{LSE-D$\downarrow$}} & \multicolumn{1}{c|}{{LSE-C$\uparrow$}}  & \multicolumn{1}{c}{{LVE$\downarrow$}} & \multicolumn{1}{c}{{FDD$\downarrow$}} 
\\ 
\hline
\multirow{3}{*}{BIWI} & {FaceFormer} & {\color[HTML]{000000} 9.101} & {\color[HTML]{000000} 9.182} & {\color[HTML]{000000} 5.3388} & {\color[HTML]{000000} 4.4103}          
\\
{} & {CodeTalker} & {9.090}   & {9.039} & {\textbf{4.8133}} & {\color[HTML]{000000} 4.1244}  
\\
{} & {Learn2Talk} & {\textbf{8.897}} & {\textbf{9.449}}  & {\color[HTML]{000000} 5.0003}           & {\textbf{3.7756}}   
\\
\hline
\hline
\multirow{3}{*}{VOCASET} & {FaceFormer} & {11.720} & {6.916} & {2.5419} & {N/A}          
\\
{} & {CodeTalker} & {12.455}   & {6.704} & {2.8708} & {N/A}  
\\
{} & {Learn2Talk} & {\textbf{10.593}} & {\textbf{9.838}}  & {\textbf{2.5148}}  & {N/A}   
\\
\hline
\end{tabular}
\caption{Methods comparison. The quantitative evaluation results over BIWI-Test-A and VOCA-Test are reported. The best score in each metric is marked with \textbf{bold}.} 
\label{tab:main_comparison}
\end{table}

\subsection{Methods Comparison}
\textbf{Competitors.} We quantitatively and qualitatively compare Learn2Talk with two state-of-the-art methods, FaceFormer \cite{FaceFormer2022} and CodeTalker \cite{CodeTalker2023}. We use their pre-trained networks on BIWI-Train and VOCA-Train for evaluation. All the three methods require conditioning on a training speaker identity during inference. For unseen subjects, we obtain the predictions of the three methods by conditioning on all training identities, and choose the best one. As FaceFormer and CodeTalker can't generate head motion, our method excludes the synthesis of head motion when making comparisons.

\begin{figure*}
    \centering
    \includegraphics[width=0.91\textwidth]{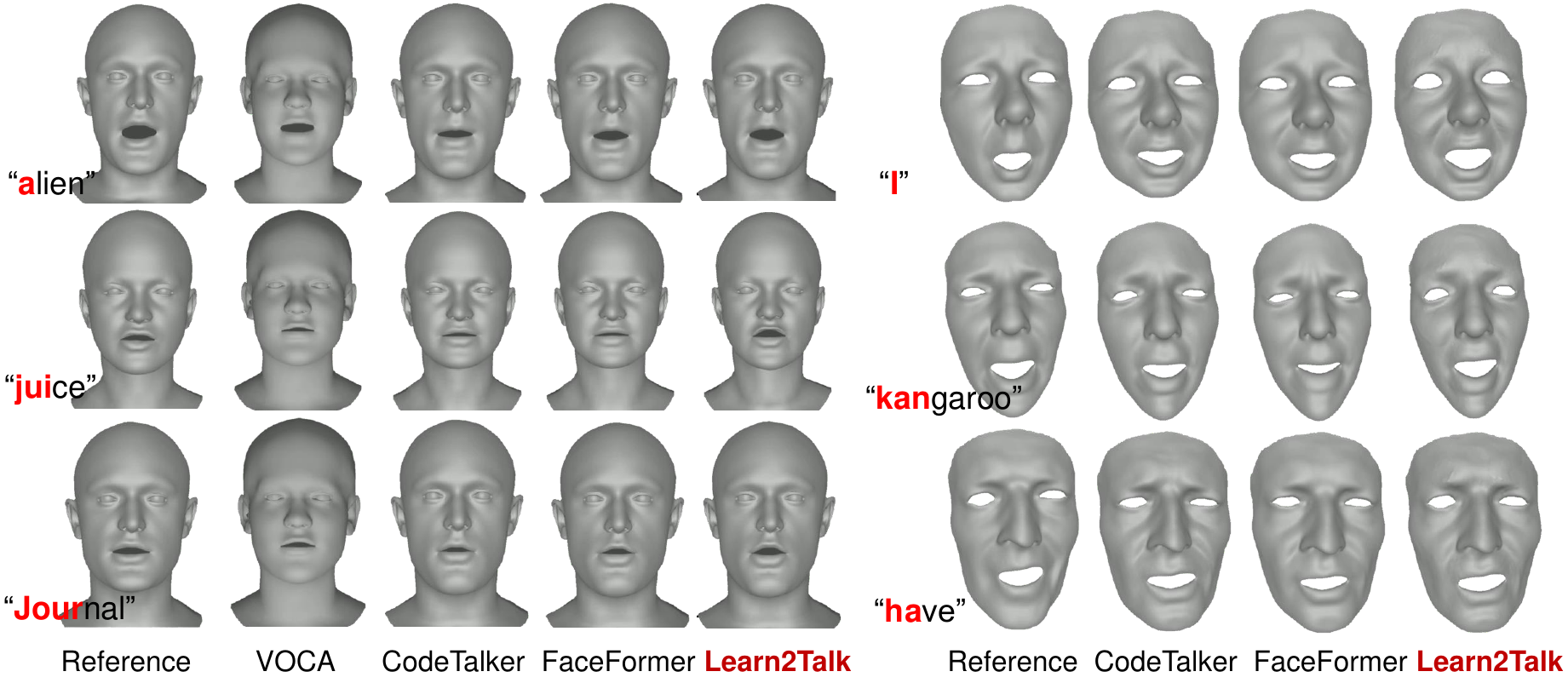}
    \caption{Visual comparisons of sampled facial motions animated by different methods on VOCA-Test (left) and BIWI-Test-A (right).}
    \label{fig:visual}
\end{figure*}

\textbf{Quantitative Evaluation.} According to the statistics in Tab. \ref{tab:main_comparison}, the proposed Learn2Talk outperforms FaceFormer and CodeTalker in the term of 3D lip-sync by a large margin, both in BIWI-Test-A and VOCA-Test, suggesting that Learn2Talk produces more accurate lip-synchronized movements than the competitors. In the metric of LVE in VOCA-Test, Learn2Talk also surpass the two competitors. When it comes LVE in BIWI-Test-A, Learn2Talk outperforms FaceFormer but slightly lags behind CodeTalker. Besides, Learn2Talk achieves better score in FDD than the competitors, indicating that Learn2Talk can enable more precise dynamic changes in upper-face. Overall, Learn2Talk has shown obvious advantages both in 3D lip-sync and 3D vertex quality, compared with the state-of-the-arts.

\textbf{Qualitative Evaluation.}
We visually compare our method with the competitors in Fig. \ref{fig:visual}. We illustrate six typical frames of synthesized facial animations that speak at specific syllables. Compared with the competitors, the lip movements produced by Learn2Talk are more accurately articulated with the speech signals and also more consistent with the reference. In the 1st row and the right part of 2nd row, Learn2Talk produces better lip-sync with proper mouth open when pronouncing diphthong /ei/, /ai/ and monophthong /$\ae$/. In the left part of 2nd and 3rd rows, Learn2Talk can generate proper pouting when pronouncing the voiced consonant in ``jui" and ``jour". In the right part of 3rd row, our mouth shape is more similar to reference when pronouncing ``ha".

\begin{figure*}
    \centering
    \includegraphics[width=0.96\textwidth]{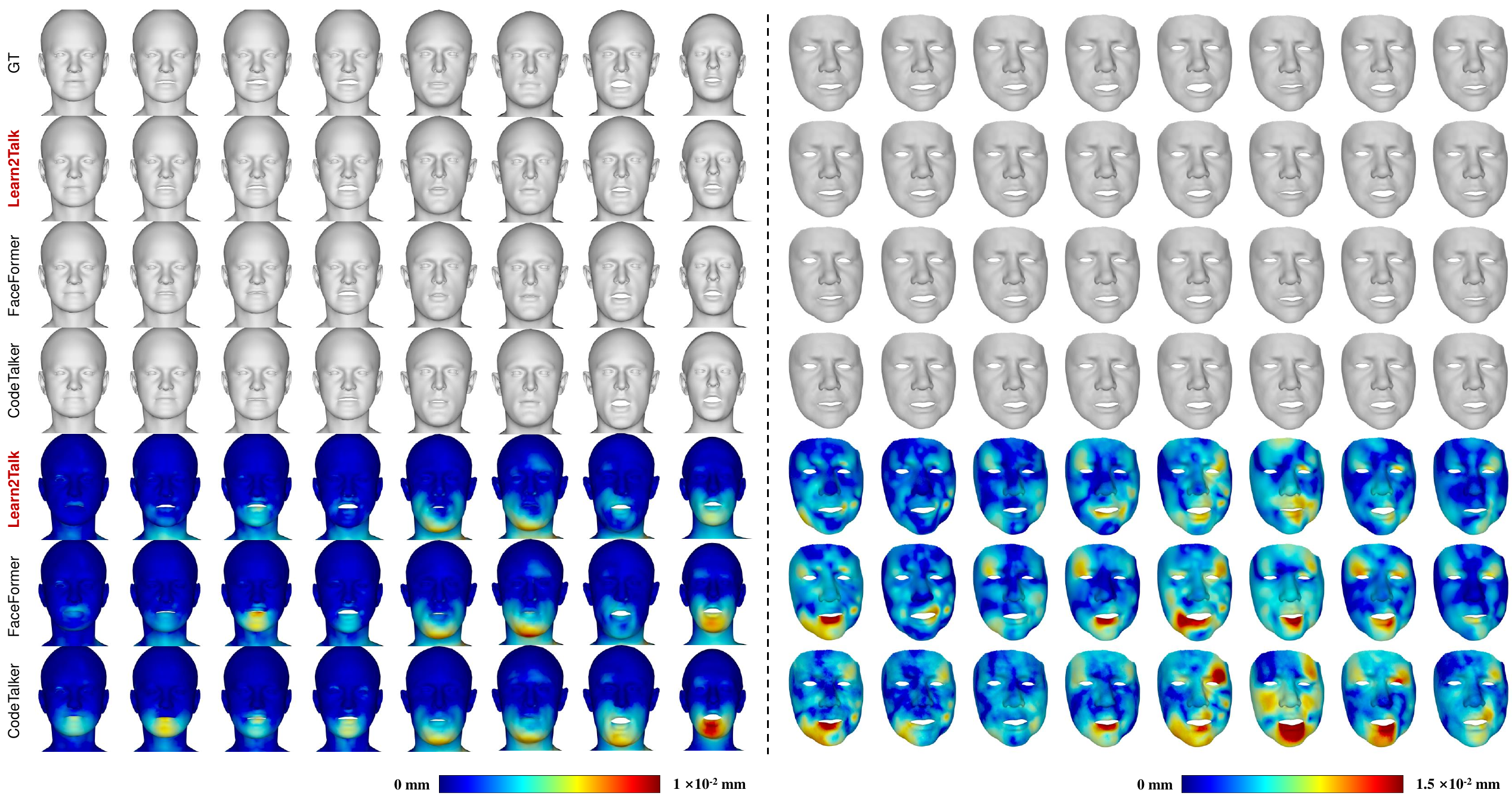}
    \caption{3D reconstruction error heatmaps of sampled facial motions predicted by different methods on VOCA-Test (left) and BIWI-Test-A (right).}
    \label{fig:heatmesh}
\end{figure*}

Fig. \ref{fig:heatmesh} shows the 3D reconstruction error heatmaps of the sampled frames, produced by the three methods. The per-vertex errors are color-coded on the reconstructed mesh for visualization. It could be observed that, our method obtains more accurate vertices both in mouth region and upper-face region.

\begin{figure}
    \centering
    \includegraphics[width=0.48\textwidth]{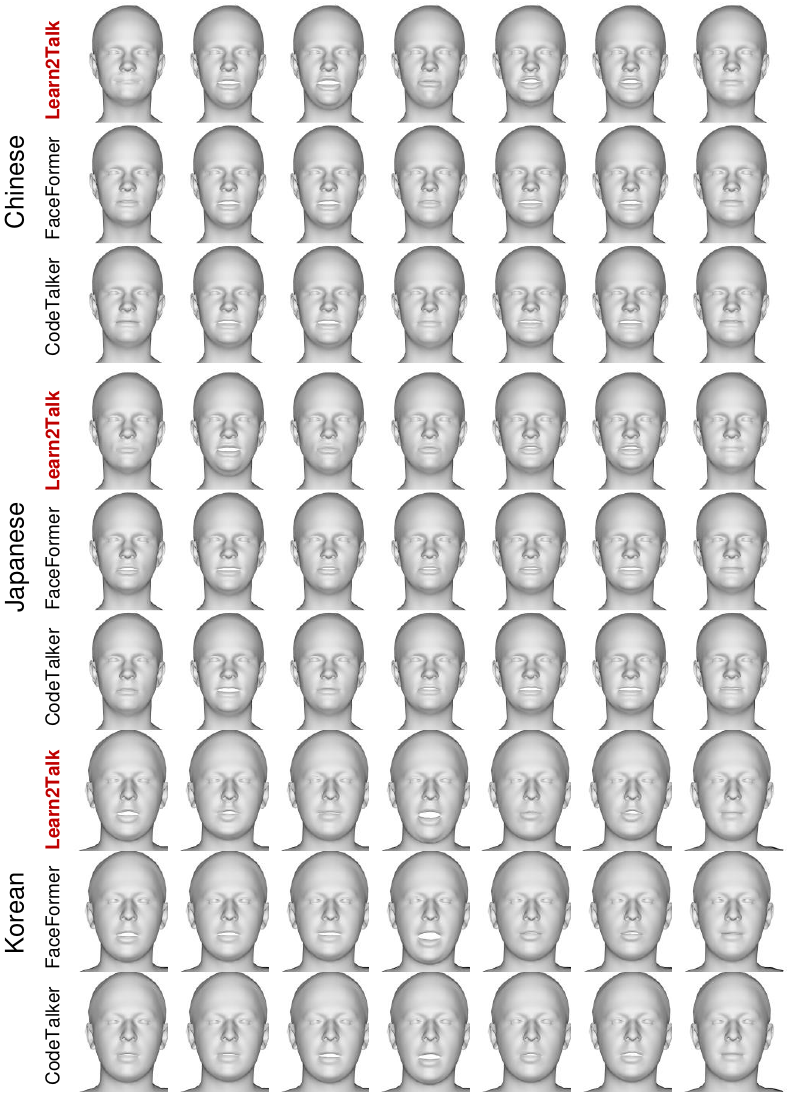}
    \caption{Visual comparisons of sampled facial motions predicted by different methods across languages, including Chinese, Japanese and Korean.}
    \label{fig:languages}
    \vspace{-10pt}
\end{figure}

Fig. \ref{fig:languages} shows the visual comparisons on other languages, including Chinese, Japanese and Korean. Although trained on English datasets, our method can generalize well to any other language. This is because that our method uses the signal characteristics of the speech audio, instead of words or linguistics. From Fig. \ref{fig:languages}, it could be observed that our method generates more expressive animation, with obvious movements in opening mouth, pouting and closing mouth. The comparisons on facial animations with all frames are available in the supplementary video.

\begin{table}[]
\centering
\begin{tabular}{l|ll}
\hline
\multicolumn{1}{c|}{} & \multicolumn{2}{c}{VOCA-Test}                      \\ 
\cline{2-3} 
{} & \multicolumn{1}{c}{Lip-sync} & \multicolumn{1}{c}{Realism}  
\\ 
\hline
{Learn2Talk vs. VOCA} & {73.57 (6.8e-7)} & {77.86 (1.6e-6)}
\\
{Learn2Talk vs. FaceFormer} & {70.71 (2.4e-3)} & {69.29 (1.0e-3)} 
\\
{Learn2Talk vs. CodeTalker} & {66.43 (4.1e-4)} & {67.14 (4.6e-5)}          
\\
\hline
\end{tabular}
\caption{User study results on VOCA-Test. The percentage of answers where Learn2Talk is preferred over the competitor is listed. The numbers in the brackets denote p-value.} 
\label{tab:userstudy}
\end{table}

\begin{table}[]
\centering
\begin{tabular}{l|l|ll|ll}
\hline
\multirow{2}{*}{Datasets} & \multirow{2}{*}{Methods} & \multicolumn{2}{c|}{3D Lip-sync}  & \multicolumn{2}{c}{3D Vertex Quality}\\ 
\cline{3-6} 
{} & {} & \multicolumn{1}{c}{{LSE-D$\downarrow$}} & \multicolumn{1}{c|}{{LSE-C$\uparrow$}}  & \multicolumn{1}{c}{{LVE$\downarrow$}} & \multicolumn{1}{c}{{FDD$\downarrow$}} 
\\ 
\hline
\multirow{3}{*}{BIWI} & {w/o sync} & {\color[HTML]{000000} 9.434} & {\color[HTML]{000000} 8.933} & {\textbf{4.6971}} & {\textbf{3.4083}}       
\\
{} & {w/o lipread} & {\textbf{8.563}} & {\textbf{9.652}} & {5.0375} & {\color[HTML]{000000} 5.1844}  
\\
{} & {full model} & {\color[HTML]{0000FF}\textbf{8.897}} & {\color[HTML]{0000FF}\textbf{9.449}}  & {\color[HTML]{0000FF}\textbf{5.0003}} & {\color[HTML]{0000FF}\textbf{3.7756}}   
\\
\hline
\hline
\multirow{3}{*}{VOCASET} & {w/o sync} & {11.717} & {6.286} & {\textbf{2.3581}} & {N/A}          
\\
{} & {w/o lipread} & {10.956}   & {\textbf{9.926}} & {2.8233} & {N/A}  
\\
{} & {full model} & {\textbf{10.593}} & {\color[HTML]{0000FF}\textbf{9.838}}  & {\color[HTML]{0000FF}\textbf{2.5148}}  & {N/A}  
\\
\hline
\end{tabular}
\caption{Ablation study. The quantitative evaluation results over BIWI-Test-A and VOCA-Test are reported. The best score in each metric is marked with \textbf{bold}, and the second best score is marked with {\color[HTML]{0000FF}\textbf{blue}}.} 
\label{tab:ablation}
\end{table}

\textbf{User Study.}
We conduct a user study to evaluate the quality of the animated faces by VOCA \cite{VOCA2019}, FaceFormer \cite{FaceFormer2022}, CodeTalker \cite{CodeTalker2023} and Learn2Talk, in the terms of perceptual lip-sync and facial realism. We adopt A/B test for each comparison, e.g. ours vs. FaceFormer, and take the random order. In each A/B test, the user watches the facial animations, listens to the audio, refers to the textual sentence, and answers the questions. About 30 A vs. B pairs are created for VOCA-Test. Each pair is judged by 15 different participants, finally yielding 450 entries. We calculate the radio of participants who prefer Learn2Taker over the competitor. The percentage of A/B testing is tabulated in Tab. \ref{tab:userstudy}, which shows that participants favor Learn2Talk over the competitors. We attribute this to the facial animation synthesized by Learn2Talk having more well-synchronized lip movements and vivid facial motions. We also conduct ANOVA tests in the user study, in which the p-value less than 0.05 indicates a statistically significant difference between our method and the competitor. In the supplementary video, we show some visual comparisons used in the user study.

\subsection{Ablation Study}
We conduct ablation study on both BIWI-Test-A and VOCA-Test to evaluate the two key designs in Learn2Talk: the lipread loss and the 3D sync loss. To remove the lipread loss in training, the modules related to the lipread loss are all removed, including differentiable renderer, lipreading network and teacher model. Similarly, to move the 3D sync loss, SyncNet3D is removed while other modules are preserved. The quantitative evaluation results are reported Tab. \ref{tab:ablation}.

\textbf{Effect of the lipread loss.} The model without lipread loss is denoted as ``w/o lipread" both in Tab. \ref{tab:ablation}. Compared with the full model, w/o lipread will introduce more errors related to 3D vertex quality while improve the metrics in lip-sync. This phenomenon indicates that lipread loss is more effective in the traditional 3D face reconstruction from audio.

\textbf{Effect of the 3D sync loss.} The model without sync loss is denoted as ``w/o sync" in Tab. \ref{tab:ablation}. Compared with the full model, w/o sync will decrease the errors related to 3D vertex quality while degenerate the metrics in lip-sync. This phenomenon clearly indicates that 3D sync loss is expert in achieving lip-sync.

\textbf{Discussion.} From the ablation study introduced above, we draw the conclusion that there is a tradeoff between the lipread loss and the 3D sync loss. These two losses are designed for different purposes, which are not harmonious in one optimization framework. The lipread loss introduces the speech articulation measured by 3D distance, while the 3D sync loss enhances the lip-sync measured by milliseconds. For example, when speaking /a:/, the lipread loss makes the mouth open as much as possible, while the 3D sync loss makes the mouth open as quickly as possible. The lipread loss yields more spatial responses, while sync loss yields only temporal responses, which makes them adversarial in training. To balance the two losses, we tune the two weighting parameters $\lambda_1, \lambda_2$ through 4 or 5 parameter combinations.

It's also worth noting that, together with the statistics reported in Tab. \ref{tab:main_comparison}, the two variants of Learn2Talk, w/o sync and w/o lipread, have more advantages than Learn2Talk itself over FaceFormer and CodeTalker respectively in 3D lip-sync and 3D vertex quality. 


\begin{table}[]
\centering
\begin{tabular}{l|l}
\hline
{Training Data} & {WER $\downarrow$}                      
\\ 
\hline
{Real Audio + Real Video (Ground Truth)} & {11.91}         
\\
{Real Audio + Synthesized Video by FaceFormer} & {13.96} 
\\
{Real Audio + Synthesized Video by CodeTalker} & {14.27} 
\\
{Real Audio + Synthesized Video by Learn2Talk} & {\textbf{13.31}} 
\\
\hline
\hline
{Synthesized Video by FaceFormer} & {56.99} 
\\
{Synthesized Video by CodeTalker} & {56.45} 
\\
{Synthesized Video by Learn2Talk} & {\textbf{50.28}} 
\\
\hline
\end{tabular}
\caption{Audio-visual speech recognition. The quantitative speech recognition results on the subset of LRS3 are reported. The lower WER, the better recognition performance. The best score in each metric is marked with \textbf{bold}.} 
\label{tab:AVSR}
\end{table}

\begin{figure*}
    \centering
    \includegraphics[width=0.98\textwidth]{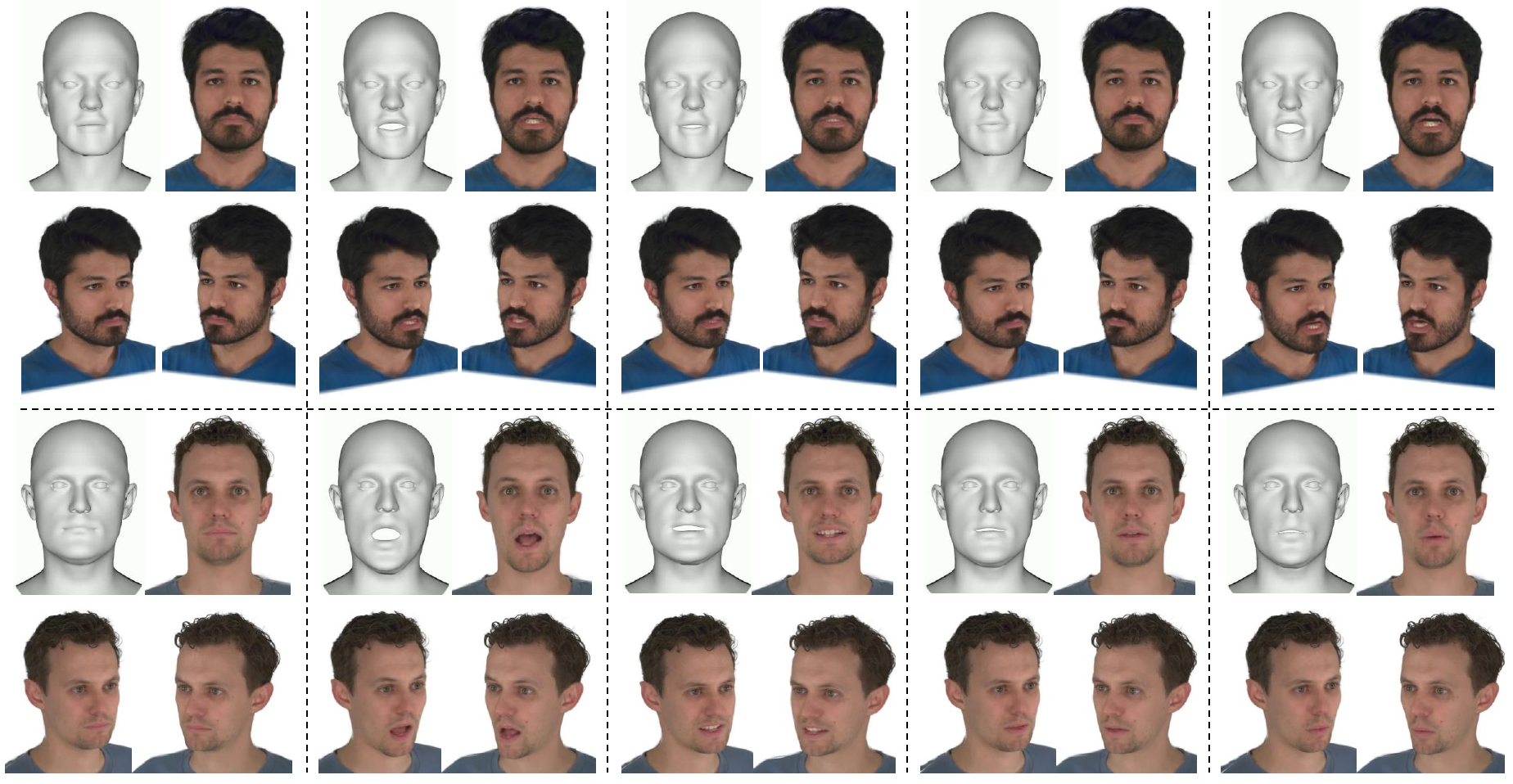}
    \caption{Visualization of the sampled frames of the speech-driven 3DGS-based head avatar animation. In each frame, we show the FLAME model generated by our method (1st row-left), and the front view (1st row-right), two side views (2nd row) of the driven avatar.}
    \label{fig:3DGS_Avatar}
\end{figure*}

\subsection{Applications}
\textbf{Audio-visual speech recognition.} The first application is to synthesize the audio-visual data for the task of the audio-visual speech recognition. We use Learn2Talk to synthesize the video clips from the audios on the subset of LRS3 \cite{LSRS3_2022}, and then construct the labeled audio-visual corpora which can be used to train and test the audio-visual speech recognition network \cite{AVSR2023}. Such synthesized audio-visual data is valuable in labeled data augmentation and personal privacy protection. The performance of the trained speech recognition networks with different type of training data is reported in Tab. \ref{tab:AVSR}. The Word Error Rate (WER) of the recognition model trained on the synthesized data by Learn2Talk (13.31\%) is closer to that of the model trained on the ground truth data (11.91\%), compared with FaceFormer (13.96\%) and CodeTalker (14.27\%). Moreover, if the ground truth audios are not involved in training and testing, the speech recognition performance degrades dramatically, but Learn2Talk (50.28\%) still performs better than FaceFormer (56.99\%) and CodeTalker (56.45\%).

To conclude, the synthesized videos from audios by our method elicit more similar speech perception with the corresponding ground-truth audios, compared with FaceFormer and CodeTalker.

\textbf{Speech-driven 3DGS-based avatar animation.} The recent 3DGS method \cite{3DGSTOG2023, 3DGSSurvey} achieves high rendering quality for novel-view synthesis with real-time performance. Instead of training the neutral network in 3D space such as NeRF \cite{NeRFECCV2020}, it optimizes the discrete geometric primitives, namely 3D Gaussians. 3DGS has already been used to reconstruct the head avatar \cite{qian2023gaussianavatars, ultraheadavatar2023} and the human body avatar \cite{HumanAvatar2023, jiang2024uv} from multiple view images. The second application of our method is that, it enables the speech-driven 3DGS-based head avatar, as shown in Fig. \ref{fig:3DGS_Avatar}. We briefly introduce how to implement this application. Firstly, we construct the head avatar of a specific subject from multiple view images by using \cite{qian2023gaussianavatars}. Then, we produce the sequence of mesh models from the driving audio by using Learn2Talk. The sequence of mesh models is further converted to the sequence of FLAME parameters \cite{FLAME2017}. The FLAME parameters in each frame are inputted to the built head avatar, finally resulting in the 3DGS animation. Our speech-driven 3DGS animation is well in-sync with the driving audio, and supports multiple view synthesis in each frame. More results of the speech-driven 3DGS animation are available in the supplementary video.

\section{Conclusion and Limitations}
Current researches on 3D and 2D talking face are almost conducted in two separate lines. To mind the gap, we propose a novel framework which learns the expertise from the field of 2D talking face to improve the capability of 3D talking face methods. By comparing to the existing state-of-the-arts, our proposed framework shows superiority in achieving lip-sync and speech perception. 

The main limitation of the proposed framework is the tradeoff between the 3D lip-sync loss and the lipread loss, which makes capabilities of our two key designs not maximized. One future work is to explore the use of multi-task learning to balance the two losses. Other limitations of our method are that, currently it doesn't include the synthesis of eye blinking, 3D gaze and emotion from driving audio. We plan to continue improving our method from the above functional aspects, aiming at achieving the highly expressive and lifelike 3D facial animation.

\bibliographystyle{IEEEtran}
\bibliography{main}

\end{document}